\documentclass[letterpaper, 10 pt, conference]{ieeeconf}  

\IEEEoverridecommandlockouts                              

\usepackage[utf8]{inputenc}
\usepackage{amsmath,array,cite,graphicx,psfrag,url}
\usepackage{amssymb,stmaryrd,soul}
\usepackage[usenames]{color}
\usepackage{Refs/des}
\usepackage{subfigure}
\usepackage{mathtools}
\usepackage{xspace}
\usepackage{algorithm}
\usepackage{algpseudocode}
\usepackage{tikz}

\def\ie{\text{i.e.}\xspace}

\def\Fig{Fig.\xspace}
\def\Figs{Figs.\xspace}

\def\taskObs{\text{task-observer}\xspace}
\def\tDual{\text{totally reciprocal}\xspace}

\newtheorem{definition}{Definition}
\newtheorem{problem}{Problem}
\newtheorem{theorem}{Theorem}

\title{\LARGE \bf
A Formal Modular Synthesis Approach for the Coordination of 3-D Robotic Construction with Multi-robots
}

\author{Marcelo Rosa$^{1}$, José E. R. Cury$^{2}$, and Fabio L. Baldissera$^{2}$ 
\thanks{$^{1}$Marcelo Rosa is with Universidade Federal de Santa Catarina, Florianópolis,
              Brazil
        {\tt\small marcelo.r.rosa@posgrad.ufsc.br}}%
\thanks{$^{2}$José E. R. Cury and  Fabio L. Baldissera are with the Department of Automation and Systems Engineering, Universidade Federal de Santa Catarina, Florianópolis, Brazil
        {\tt\small {jose.cury, baldissera}@ufsc.br}}%
}

\begin{document}

\maketitle
\thispagestyle{empty}
\pagestyle{empty}

\begin{abstract}
In this paper, we deal with the problem of coordinating multiple robots to build 3-D structures.
This problem consists of a set of mobile robots that interact with each other in order to autonomously build a predefined 3-D structure.
Our approach is based on Supervisory Control Theory and it allows us to synthesize from models that represent a single robot and the target structure a correct-by-construction reactive controller, called \emph{supervisor}.
When this supervisor is replicated for the other robots, then the target structure can be completed by all robots.  
\end{abstract}

\section{INTRODUCTION}

Nature has served as a source of inspiration for the development of new technologies in the fields of engineering, computer science, and robotics.
In particular, we can mention the collective construction systems where a set of individuals with limited abilities organize themselves to perform a complex task that would be difficult or even unfeasible for a single individual.
Among the many examples that can be observed in nature, we can mention the construction of dams by families of beavers, beehives by swarms of bees, anthills by ants, and termite mounds by termites.
Despite the differences among such systems, they all share one common feature: the absence of centralized control \cite{hansell:2007}.
Therefore, it can be said that coordination in such cooperative systems is inherently decentralized and scalable in terms of the number of individuals.

Collective construction systems have inspired the development of autonomous robotic systems that, despite being composed of a set of simple robots, can build a wide range of structures, such as walls and bridges \cite{Stewart:2006, Werfel:2006, Petersen:2012, Rubenstein:2014, Soleymani:2015, Soorati:2016, Saldana:2017}.
Although these systems are still limited to working with relatively small structures, in controlled environments, and with materials that simulate the real ones, it is expected that they may be a viable alternative for the construction of temporary shelters or containment structures after some natural disaster \cite{Petersen:2019}.

In the context of robotic construction, there is a problem of coordination that has stimulated new research
\cite{Werfel:2014, Deng:2019, MR:WODES2020} and international competitions \cite{mbzirc:2020}. 
This problem basically consists of a set of mobile robots in a world of bricks whose a common objective is the construction of a given 3-D structure.
The structure is composed from stacks of blocks, which must be carried and rearranged by the robots.
The objective in this problem is to determine a decentralized robot coordination strategy that is not only scalable in relation to the number of robots but also displays \emph{permissiveness} (i.e., the local controllers implemented in each robot act reactively in order to prune only those actions that disobey some specification).


In this paper\footnote{This paper is an application from results introduced in \cite{MR:WODES2022}.}, we propose a modular controller synthesis process based on Supervisory Control Theory (SCT)\cite{RW:1989} to solve the 3-D multi-robot construction problem. 
In our approach, the behavior of a single robot and the target 3-D structure are modeled by automata.
Then, these automata are used to synthesize a reactive controller, called \emph{supervisor}, that satisfies a set of properties. 
This set of properties ensures that if we replicate the synthesized supervisor for the other robots, then the target structure can be collectively completed by all robots.
In our control framework, each robot has its own supervisor that observes the actions executed by its robot and the addition of bricks performed by the other robots and in response to each observed event provides a set of enabled next possible actions.
Then, the robot can execute any one among those actions.

The contributions of this paper are: %
(1) a systematic and formal way to model the robot and the structure; %
(2) a modular method for synthesizing reactive controllers for the robots, which ensures that the target structure is completed by all robots working together.
This method allows us to synthesize a correct-by-construction supervisor for a robot from the models that represent the robot and the structure; and
(3) a decentralized reactive control solution that provides multiple paths for the robots to construct the target structure.

The remainder of the paper is organized as follows: Section~\ref{sec:ProblemState} introduces the 3-D multi-robot construction problem.
Section~\ref{sec:SO} presents a solution overview.
Section~\ref{sec:solving} describes the details of our method and states the conditions that guarantee that the 3-D multi-robot construction problem is solved.
Finally, Section~\ref{sec:conc} concludes the paper.

\section{Problem Statement}\label{sec:ProblemState}

In this section we formalize the 3-D robotic construction problem.
In this problem, a set of mobile robots must build a 3-D structure composed of only one type of brick. 
Before we state the problem, a few definitions are needed. 

\begin{figure}[ht!]
    \centering
    \subfigure[{Structure \St}]{
        \includegraphics[scale = 0.8]{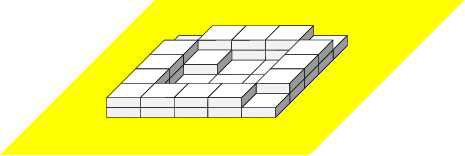}
        \label{img:3D}
    }
    
    \subfigure[{function $h_{\St}$}]{
        \psfrag{00}{\tiny $(0,0)$}
        \psfrag{xy}{\tiny $(x,y)$}
        \psfrag{x}{\tiny$x$}
        \psfrag{y}{\tiny $y$}
        \includegraphics[scale = 0.6]{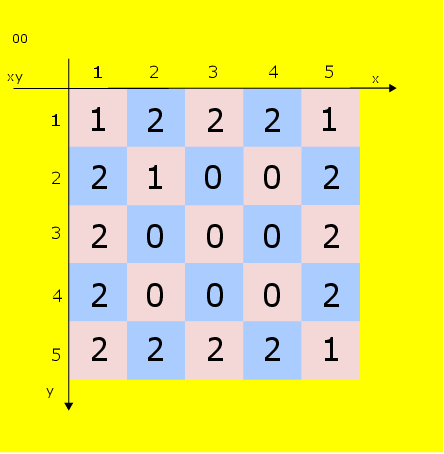}
        \label{img:2D}
    }
    \caption{ %
        \subref{img:3D} Example of a 3-D structure composed of equal bricks.
		\subref{img:2D} Alternative representation of the 3-D structure in \subref{img:3D} by function $h_{\St}$, where each cell $(x \bcom y)$ in two-dimensional grid displays the number of bricks stacked on $(x \bcom y)$.%
		The pair $(0,0)$ represents the region surrounding the construction site. 
		The indexes $x$ and $y$ in $(x \bcom y)$ denote the column and line in the grid, respectively. 
		In this paper, we enumerate the cells according to the convention shown here, that is, indices increase top to bottom and left to right. 
    }
    \label{img:strt}
\end{figure}

\begin{definition}
	A 3-D structure \St is modeled by a function $h_{\St}:\mathbb{N}\times\mathbb{N}\to \mathbb{N}$ defined on a bounded and closed domain $\Do$, also called the \emph{construction site}.  
\end{definition}
The domain $\Do \subseteq \mathbb{N}\times\mathbb{N}$ of $h_{\St}$ is interpreted as the grid where the structure must be built, whereas $h_{\St}(x,y)$ denotes the height of a particular cell $(x,y)$, i.e., the number of piled bricks it contains.
Note that a 3-D structure is always delimited in space and that all bricks are assumed to be equal.
\Fig~\ref{img:strt} illustrates the mathematical representation of  a 3D-structure. 

We use $\Do'$ to denote the set of all cells $(x,y) \in \Do$, where there is more than one brick in the structure \St, i.e. $h_{\St}(x, y) \ge 1$.

At the construction site there is a special set of locations $\IO \subseteq \Do$ which correspond to locations through which the robots can either enter or leave the construction site.
Furthermore, the region surrounding the construction site is identified by the pair $(0 \bcom 0)$ and $h_S(0 \bcom 0)$ is always zero. 

Two cells $(x,y)$ and $(x',y')$ of the domain $\Do$ of $h_S$ are said to be neighbors if $|x - x'| + | y - y'| = 1$.
For a cell $(x,y)$, we denote by $\N[{x,y}]$ the set of all its neighboring cells. 
We further assume that the location $(0,0)$ is a neighbor of all enter and exit locations, \ie, $\N[0,0] = \IO$ and $(0,0) \in \N[x,y]$, for all $(x,y) \in \IO$.

Now, we introduce the construction robots that are employed to build the structure according to their abilities. 
A construction robot is capable of:
	\begin{itemize}
		\item picking up one brick at a time of buffer;
		\item unloading a brick at $(x,y)$ whenever the robot is at a neighbor cell $(x', y') \in \N[x,y]$ of the same height as $(x,y)$, that is, $h_{\St}(x,y) = h_{S}(x',y')$; 
		\label{cr:unload}
		\item moving across neighboring cells $(x,y)$ and $(x',y')$, one step at a time, and such that  $|h_{St}(x,y)-h_{\St}(x',y')| \leq 1$, \ie, the robots can not move through steps higher than one brick.  \label{cr:move}
	\end{itemize}

\begin{figure}[ht!]
	\centering
	\includegraphics[scale= 0.8]{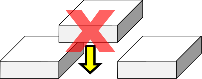}
	\caption{A robot cannot unload a brick in the middle of other bricks.}
	\label{img:exSpecU}
\end{figure}

Given the fact that inserting a brick onto a one-brick-wide space is mechanically difficult and requires a high precision \cite{Werfel:2014}, then, a robot cannot unload a brick on a cell $(x,y)$ whose height $h_{S_c}(x,y)$ satisfies either:
	a) $h_{\St}(x,y) < h_{\St}(x-1,y)$ and $h_{\St}(x,y) < h_{S_c}(x+1,y)$; or b) $h_{\St}(x,y) < h_{\St}(x,y-1)$ and $h_{\St}(x,y) < h_{\St}(x,y+1)$. 
	Such specification is illustrated in \Fig~\ref{img:exSpecU}. 

In our approach, we assume that: 
\begin{enumerate}
	\item[(A1)] there is an infinite buffer of bricks outside the construction site;
	\item[(A2)] the robot actions - move, pick up and place a brick - are controllable and observable, that is, the robot control system can prevent these actions from occurring as well as it can detect when they occur; 
	\item[(A3)] there is a communication channel that allows the robots to exchange messages with each other as follows:
	\begin{enumerate}
	    \item whenever a robot unloads a brick on a grid position $(x,y)$, it broadcasts this information to the other robots;
	    \item when a robot intends to unload a brick on $(x,y)$, it requests permission from other robots. 
	\end{enumerate}
	\item[(A4)] collision avoidance is implemented by a lower level controller, \ie, a robot is equipped with a mechanism that prevents it to enter a cell occupied by any other robot or put a brick on such cell.
    In general, the grid size is much greater than the number of robots working, so there is a considerable space for the robots to maneuver to avoid collision.
\end{enumerate}

Remark that, those assumptions do not impose a priori any constraints on the shape of the structure.
This fact leads to a wider range of possible structures to be built, when compared to other approaches in the literature as we argue next.
For example, the structure \St shown in \Fig~\ref{img:strt} is feasible, but it cannot be built by using the approaches proposed in \cite{Werfel:2014, Deng:2019}, where they assume that the final structure must be \emph{traversable}, \ie, for each location there must be a \emph{closed path}\footnote{A closed path is a sequence of locations $ l_0 \bcom l_1 \bcom \cdots \bcom l_n \bcom l_0$ that begin and end at $l_0$ such that consecutive locations are neighbors.} %
starting from $(0 \bcom 0)$ that reaches it, and such that no location is visited twice except $(0 \bcom 0)$.   

To coordinate the actions of the robots, it is considered that each robot has a reactive controller, called \emph{supervisor}, that takes as input actions performed by the robot itself and changes in the structure caused by other robots (addition of a brick) and then it determines which are the next enabled set of actions for the robot under its control.
This set is updated after the occurrence of each new input.
Then, the robot can choose any of the actions from this set to perform, a feature associated with a permissive supervisor.
The permissive nature of the supervisor allows specific criteria to be used to determine which action, among those enabled by the supervisor, will be performed by the robot.
Regardless of the criteria adopted in decision making, a supervisor ensures that the structure is completed without any control constraints being violated.
In this paper, we focus on the synthesis of the supervisor and do not approach these (low level) criteria.

Finally, we state the 3-D multi-robot  construction problem as follows.

\begin{problem}\label{prob:main}
	Given a target 3-D structure \St, a set of construction robots and assumptions (A1)-(A4), find, if it exists, a supervisor for each robot such that: 
	\begin{enumerate}
		\item[a)] \St is obtained from an initial configuration $\St[0]$, with $h_{\St[0]}(x \bcom y)=0$ for all $(x \bcom y) \in \Do$; 
		\item[b)] all robots satisfy their behavioral constraints; and
		\item[c)] all robots end up outside the construction site. 
	\end{enumerate}
\end{problem}


\section{Solution Overview}\label{sec:SO}

Our approach for solving Problem 1 is composed of the following steps:

\begin{enumerate}
	\item \emph{Structures}: obtain a nonblocking finite automaton  $\T = \etuple[{\T}]$ whose states correspond to all feasible intermediate structures bounded by the target structure \St, and state transitions correspond to the possible adding brick actions performed by some robot, events $\evtT{x,y}{} \in \Alf[{\T}]$, with $x \in X$ and $y \in Y$.
	Event $\evtT{x,y}{}$ means that some robot added a brick on cell $(x,y)$.
	For each state $\St[k]$ from $\T$ the addition of an extra brick on cell $(x,y)$, events $\evtT{x,y}{}$, is only possible if
	\begin{enumerate}
		\item there exists at least one neighboring cell $(x',y') \in \N[x,y]$ with the same height as $(x,y)$;
		\item for any two neighboring cells $(x',y') \bcom (x'',y'') \in \N[x,y]$, such that either $x' = x'' = x$ or $y' = y''= y$, at least one of them must be at the same height or below of $(x,y)$;
		\item cell $(x,y)$ has not reached its desired height.
	\end{enumerate}
	Remark that the aforementioned conditions consider only the state of the structure itself and, therefore, $\T$ model is independent of both the number of robots and their status (current position and load).
	\Fig~\ref{img:strtAut} shows part of the automaton \T for structure \St from \Fig~\ref{img:strt}.
	
	\begin{figure}[ht!]
	    \centering
	    \psfrag{t11}{\small $\evtT{1,1}{}$}
	    \psfrag{t12}{\small $\evtT{1,2}{}$}
	    \psfrag{t55}{\small $\evtT{5,5}{}$}
	    \includegraphics[width=\linewidth]{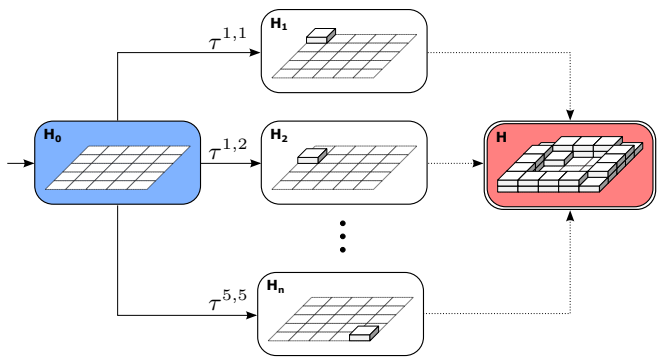}
	    \caption{Part of the automaton \T for structure \St from \Fig~\ref{img:strt}.
	    In this paper, we adopted by convention that the initial state (blue state) is indicated on figure by an empty single incoming arrow and marked states (red state) by double lines.
	    The dashed arrows encapsulate sequences of states and transitions.
	    }
	    \label{img:strtAut}
	\end{figure}
	\item \emph{Robots}: construct a finite automaton $\R[i] = \etuple[{\R[i]}]$, with $i \in \I = \{1, \ldots, n\}$, that models the behavior of the robot \R[i] according to its primitive actions and hardware limitations such as climbing, adding and load restrictions.
	\Alf[{\R[i]}] is the union of the disjoint subsets \Alftask[{\R[i]}]  and \Alfloc[{\R[i]}], containing the \emph{unload} and the \emph{local} events, respectively.
	The set of unload events is given by $\Alftask[{\R[i]}] = \{ \evtT{x,y}{i} \mid (x,y) \in \Do' \}$, where $\evtT{x,y}{i}$ means that robot \R[i] unloaded a brick on $(x,y)$.
	Conversely, \Alfloc[{\R[i]}] contains the events that represent the moving and loading actions of \R[i].
	%
	\item \emph{Modular synthesis}: compute the supervisor $\S[i] = \Synt(\T[i] \| \R[i])$, with $i \in \I$, where $\T[i] \| \R[i]$ stands for the \emph{synchronous product}\footnote{Shared events between two automata are synchronized in lock-step, while other events are interleaved \cite{cassandras:2008}.} of the automata \T[i] and \R[i], and \T[i] is the unfolded local version of \T for robot \R[i], in which each transition labeled with an event \evtT{x,y}{} is replaced by a set of transitions labeled with events $\evtT{x,y}{i}$ and $\evtT{x,y}{\Other}$.
	\item \emph{Replicating supervisors}: replicate $\S[i]$ for all \S[j], with $j \ne i$. 
	To do this, simply replace the index $i$ with $j$ from events $\evtT{x,y}{i} \in \Alftask[{\R[i]}]$ and $\sigma_{i} \in \Alfloc[{\R[i]}]$.
\end{enumerate}

In our modular synthesis, \Synt is an iterative procedure that removes states from the automaton $\T[i] \|\R[i]$ that are not co-accessible, that is, states that do not lead to a marked state.
Moreover, \Synt ensures two extra properties: \emph{\taskObs} and \emph{\tDual}.
In practice, these two properties assure that the joint controlled behavior of the robots is nonblocking and leads to the target structure \St, as it will be shown in section~\ref{sec:solving}.
As a result, \Synt provides a supervisor represented by an automaton \S[i] that runs concurrently to \R[i] and it acts by disabling the eligible events in \R[i] that are not currently eligible in \S[i], after the occurrence of a sequence of events. 
Moreover, \S[i] may disable unload events from other robots (events \evtT{x,y}{\Other}).
For an event \evtT{x,y}{\Other} disabled by \S[i] to be effectively disabled, it is necessary that
this disablement could be transmitted to the other supervisors.
The marked states of \S[i] indicates that structure \St was completed and robot $i$ is at $(0,0)$.

\section{Solving the 3-D multi-robot construction problem}\label{sec:solving}

In this section we will show that if the supervisor \S[i] designed for robot \R[i] satisfies the nonblocking, \taskObs and \tDual properties, then the joint action of supervisors is nonblocking, that is, the target structure is collectively completed by all robots and they all return out of the structure.
For that, we present how \T[i] and \R[i] models can be constructed and used to synthesize a supervisor \S[i].

\subsection{Structures}\label{sec:structure}

The purpose here is to build the automaton \T[i].
For that, we employ two \emph{extended automata with variables} \cite{Chen:2000,Fabian:2007, Ouedraogo:2011}, \G[1] and \G[2].
\G[1] stores the information about the structure under construction and describes the possible changes in the structure caused by any robot, whereas $\G[2]$ defines the target structure. 
We highlight that \G[1] and \G[2] are templates that depend on the grid size and the final heights of each cell $(x,y)$.
Thus, \T[i] may obtained through $\trim(\G[1] \| \G[2])$, where \trim is a function that remove all non accessible or co-accessible states in $(\G[1] \| \G[2])$ \cite{cassandras:2008}.

\begin{figure}[ht!]
	\centering
	\begin{tikzpicture}[scale=0.2]
        \tikzstyle{every node}+=[inner sep=0pt]
        \draw [black] (2.3,-4.8) circle (2);
        \draw [black] (2.3,-4.8) circle (1.4);
        \draw (23.5,-6.8) node {$h_{x,y} := h_{x,y} + 1$};
        \draw (20,-4.8) node {$g_a \land g_b$};
        \draw [black] (2.3,-0.2) -- (2.3,-2.8);
        \fill [black] (2.3,-2.8) -- (2.8,-2) -- (1.8,-2);
         \draw [black] (4.086,-3.918) arc (144:-144:1.5);
        \draw (7.3,-4.8) node [right] {$\evtT{x,y}{i} \bcom \evtT{x,y}{\Other} :$};
        \fill [black] (4.09,-5.68) -- (4.44,-6.56) -- (5.03,-5.75);
    \end{tikzpicture}
	\caption{%
		\G[1] model to update of the variables $h_{x,y}$.
		Guard $g_a$ is given by the Boolean-valued function $\exists (x',y') \in \N[x,y] \mid h_{x',y'} = h_{x,y}$, whereas $g_b$ is given by $\exists (x',y') \bcom (x'',y'') \in \N[x,y] \mid (h_{x,y} \ge h_{x',y'} \lor h_{x,y} \ge h_{x'',y''} ) \land (x' = x'' = x \oplus x' = x'' = y)$.
		A transition with whether $\evtT{x,y}{i}$ or $\evtT{x,y}{\Other}$ can only occur if $g_a \land g_b$ is evaluated to \emph{true}, and when it occurs, \G[1] changes of state while the value of $h_{x,y}$ is increased by 1. 
	}
	\label{img:Gaxy}
\end{figure}
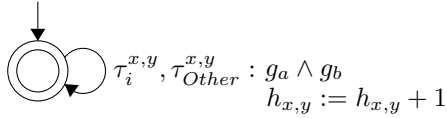

As the goal of $\G[1]$ is to maintain the information about the current configuration of the structure, we define a variable $h_{x,y}$ for each location $(x,y)$ in the grid, which stores the current number of bricks at $(x,y)$. 
The updating of variables $h_{x,y}$ is associated with occurrence of events $\evtT{x,y}{i}$ and $\evtT{x,y}{\Other}$, so that whenever events $\evtT{x,y}{i}$ or $\evtT{x,y}{\Other}$ occur the current value of $h_{ij}$ is increased by $1$.
As described in Section~\ref{sec:SO}, the addition of an extra brick on cell $(x,y)$ can only occur under certain conditions.
First, we model the conditions a) and b), respectively, through guards    
\begin{itemize}
	\item $g_a = \exists (x',y') \in \N[x,y] \mid h_{x',y'} = h_{x,y}$ and
	\item $g_b = \exists (x',y') \bcom (x'',y'') \in \N[x,y] \mid (h_{x,y} \ge h_{x',y'} \lor h_{x,y} \ge h_{x'',y''} ) \land (x' = x'' = x \oplus x' = x'' = x)$,
\end{itemize}
where $g_a$ refers to the existence of at least one neighboring cell $(x',y') \in \N[x,y]$ with the same height as $(x,y)$, whereas $g_b$ that at least one of two neighboring cells that are in the same row or column that $(x,y)$ must have its height equal to or less than $h_{x,y}$.
\Fig~\ref{img:Gaxy} shows $\G[1]$ that models the update of the variables $h_{x,y}$ conditioned to guards $g_a$ and $g_b$.

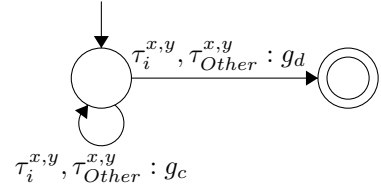
\begin{figure}[ht!]
    \centering
    \begin{tikzpicture}[scale=0.2]
        \tikzstyle{every node}+=[inner sep=0pt]
        \draw [black] (2.2,-5.3) circle (2);
        \draw [black] (18.6,-5.3) circle (2);
        \draw [black] (18.6,-5.3) circle (1.4);
        \draw (10,-3.8) node {$\evtT{x,y}{i}, \evtT{x,y}{\Other}:g_d$};
        \draw [black] (2.2,-0.2) -- (2.2,-3.3);
        \fill [black] (2.2,-3.3) -- (2.7,-2.5) -- (1.7,-2.5);
        \draw [black] (4.2,-5.3) -- (16.6,-5.3);
        \fill [black] (16.6,-5.3) -- (15.8,-4.8) -- (15.8,-5.8);
        \draw [black] (3.082,-7.086) arc (54:-234:1.5);
        \draw (2.2,-10.3) node [below] {$\evtT{x,y}{i}, \evtT{x,y}{\Other}:g_c$};
        \fill [black] (1.32,-7.09) -- (0.44,-7.44) -- (1.25,-8.03);
    \end{tikzpicture}
    \caption{ %
        Templates for $\G_{x,y}^{2}$, with $g_c := h_{x,y} < h_{\St}(x,y) - 1$ and $g_d := h_{x,y} = h_{\St}(x,y) - 1$.
        Note that, the transition to marked state in $\G_{x,y}^2$ is enabled whenever the height at $(x \bcom y)$ is equal to $h_{\St}(i \bcom j)-1$. 
        In this situation, if the robot $\R[i]$ or another robot stack one more brick on $(x,y)$, events $\evtT{x,y}{i}$ and $\evtT{x,y}{\Other}$, then the target height at $(x,y)$ is achieved and $\G_{x,y}^2$ goes to the marked state.
    }
    \label{img:specSf}
\end{figure}

It remains to model $\G[2] = \Sync_{(x,y) \in \Do'} \G_{x,y}^2$ that defines the target 3-D structure $\St$.
For this, we exploit the idea of completed tasks associated with marked states to define $\St$ so that each location $(x \bcom y)$ has an extended automaton $\G_{x,y}^2$.
Thereby, whenever $(x \bcom y)$ reaches the target height $h_{\St}( x \bcom y)$, then its respective extended automaton $\G_{x,y}^2$ will be in a marked state.   
\Fig~\ref{img:specSf} shows $\G_{x,y}^2$.

We note that states marking of \G[1] and \G[2] has been chosen so that when $\St$ is completed, then both \G[1] and \G[2] are in a marked state.
Moreover, each state from $\T[i] = \trim(\G[1] \| \G[2])$ can be associated with an intermediate structure \St[k].

\subsection{Robots}\label{sec:robot}

\begin{figure}[ht!]
    \centering
    \begin{tikzpicture}[scale=0.2]
        \tikzstyle{every node}+=[inner sep=0pt]
        \draw [black] (2.2,-4.9) circle (2);
        \draw (2.2,-4.9) node {$0$};
        \draw [black] (2.2,-4.9) circle (1.4);
        \draw [black] (16.5,-4.9) circle (2);
        \draw (16.5,-4.9) node {$1$};
        \draw [black] (16.5,-4.9) circle (1.4);
        \draw [black] (2.2,-0.2) -- (2.2,-2.9);
        \fill [black] (2.2,-2.9) -- (2.7,-2.1) -- (1.7,-2.1);
        \draw [black] (3.836,-3.755) arc (119.8234:60.1766:11.087);
        \fill [black] (14.86,-3.75) -- (14.42,-2.92) -- (13.92,-3.79);
        \draw (9.35,-1.79) node [above] {$p_i$};
        \draw [black] (14.841,-6.013) arc (-61.16648:-118.83352:11.387);
        \fill [black] (3.86,-6.01) -- (4.32,-6.84) -- (4.8,-5.96);
        \draw (10,-7.4) node [below] {$\evtT{1,1}{i}, \ldots, \evtT{n_x, n_y}{i}$};
    \end{tikzpicture}
    
    \caption{
        \G[4] models the fact that the robot $i$ can only unload a brick on a given cell $(x,y)$, event $\evtT{x,y}{i}$, after having picked a brick from the buffer, event $p$.
		Furthermore, it models loading capacity of the robot, since after event $p$ occurs it only becomes possible again if any \evtT{x,y}{i} occurs.}
    \label{img:Gload}
\end{figure}
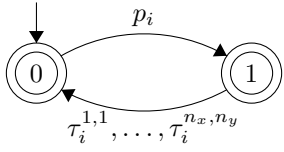

Now, we consider how to model the behavior of a construction robot.
Similar to \T[i], \R[i] may be obtained in a modular way, by the composition of extended automata \G[3], \ldots, \G[7] that represent particular aspects from its behavior. 
In our approach, we consider that the  robot moves by advancing to a single cell each time in one of the following directions: east (event $e_i$), west (event $w_i$), north (event $n_i$) and south (event $s_i$).
In addition, we define extra moves: move inside the construction site through one of entry cells $(x,y) \in \IO$ (events $in_i^{x,y}$) and outside of it (event $out_i$).
To store the current position of the robot in the grid we use the variables $\x$ (for the column) and $\y$ (for the line), whose domains are $\DOM[\x] = \{0 \bcom \cdots \bcom n_x\}$ and $\DOM[\y] = \{0 \bcom \cdots \bcom n_y\}$, respectively.
The robot starts outside the grid, \ie, at location $(0 \bcom 0)$, and the initial value of both variables $\x$ and $\y$ is set to $0$.
After starting, the possible robot action is to pick up a brick from the buffer (event $p_i$) and then move to one of entry cells $(x,y) \in \IO$ (event $in_{x,y}$).
When the robot is loaded at cell $(\x \bcom \y)$, it can put a brick on one of its neighboring cells $(x \bcom y) \in \N[\x, \y]$ (event $\evtT{x,y}{i}$). 
Accordingly, the event sets associated with the aforementioned robot actions are $\Alftask[i] = \{ \evtT{x,y}{i} \mid (x,y) \in \Do'\}$ and $\Alfloc[i] = \{e_i \bcom w_i \bcom n_i \bcom s_i \bcom in_i^{x,y} \bcom out_i \bcom  p_i\}$, where $in_i^{x,y}$ is instantiated for each cell $(x,y) \in \IO$.
The robot navigation and its loading are modeled by $\G_3$ as shown in \Fig~\ref{img:Gmove}, whereas $\G_4$ in \Fig~\ref{img:Gload} models the fact that the robot can stack a brick on $(x \bcom y)$, event $\evtT{x,y}{i}$, only after having picked it up from the buffer situated at $(0 \bcom 0)$ (event $p_i$).

\begin{figure}[ht!]
    \centering
    \begin{tikzpicture}[scale=0.2]
        \tikzstyle{every node}+=[inner sep=0pt]
            \draw [black] (6.3,-8) circle (2);
            \draw (6.3,-8) node {$q_0$};
            \draw [black] (20.6,-8) circle (2);
            \draw (20.6,-8) node {$q_1$};
            \draw [black] (20.6,-8) circle (1.4);
            \draw (12.3,-9.8) node {$out: \x \coloneqq 0$};
            \draw (14.3,-11.4) node {$\y \coloneqq 0$};
            \draw (0.7,-8) node {$p_i$};
            \draw (12.3,-2.8) node {$in_i^{x,y}: \x \coloneqq x$};
            \draw (14.8,-4.2) node {$\y \coloneqq y$ };
            \draw (22.7,-2.4) node {$w_i: \x \coloneqq -1$};
            \draw (23.5,-0.6) node {$\x > 1$};
            \draw (28.3,-4.9) node {$e_i: \x < n_x$};
            \draw (30.8,-6.5) node {$\x \coloneqq \x + 1$};
            \draw (22.7,-13.4) node {$s_i: \y < n_y$};
            \draw (25.3,-15.) node {$\y \coloneqq \y + 1$};
            \draw (28.1,-9.8) node {$n_i: \y > 1$};
            \draw (31.3,-11.8) node {$\y \coloneqq \y - 1$};
            \draw [black] (6.3,-3.3) -- (6.3,-6);
            \fill [black] (6.3,-6) -- (6.8,-5.2) -- (5.8,-5.2);
            \draw [black] (7.936,-6.855) arc (119.8234:60.1766:11.087);
            \fill [black] (18.96,-6.85) -- (18.52,-6.02) -- (18.02,-6.89);
            \draw [black] (20.76,-9.986) arc (32.34133:-255.65867:1.5);
            \fill [black] (19.12,-9.33) -- (18.18,-9.34) -- (18.71,-10.19);
            \draw [black] (18.6,-8) -- (8.3,-8);
            \fill [black] (8.3,-8) -- (9.1,-8.5) -- (9.1,-7.5);
            \draw [black] (22.536,-8.471) arc (104.07108:-183.92892:1.5);
            \fill [black] (21.4,-9.82) -- (21.11,-10.72) -- (22.08,-10.48);
            \draw [black] (21.607,-6.281) arc (177.36982:-110.63018:1.5);
            \fill [black] (22.58,-7.75) -- (23.35,-8.29) -- (23.4,-7.29);
            \draw [black] (19.238,-6.546) arc (250.86727:-37.13273:1.5);
            \fill [black] (20.93,-6.03) -- (21.66,-5.44) -- (20.72,-5.11);
            \draw [black] (4.514,-8.882) arc (-36:-324:1.5);
            \fill [black] (4.51,-7.12) -- (4.16,-6.24) -- (3.57,-7.05);
    \end{tikzpicture}
    \caption{%
        \G[3] models the navigation and loading of robot \R[i].
		State \state[0] and \state[1] have the semantics `robot outside the grid' and `robot inside the grid', respectively.
		State \state[0] is chosen to be a marked state.
	    Take, as an example, the transition associated with the event $e_i$ (move east): it can only occur when the robot is on a cell that is not located at the lateral border of the grid (guard $\x < n_x$); upon its occurrence, the $\x$ variable has its value incremented by one (action $\x\coloneqq \x+1$).
		Guard $g_1$ for the event `out' is given by the atomic proposition $(\x \bcom \y)\in \IO$.
    }
    \label{img:Gmove}
\end{figure}
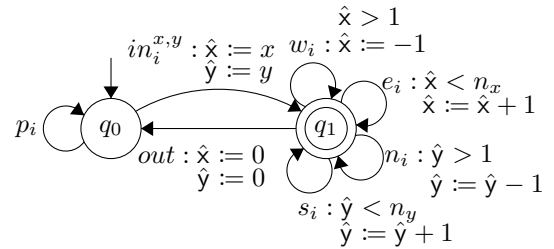

\begin{figure}[ht!]
    \centering
    \subfigure[{\G[5]}]{
        \begin{tikzpicture}[scale=0.2]
            \tikzstyle{every node}+=[inner sep=0pt]
            \draw [black] (6.8,-6.4) circle (2);
            \draw [black] (6.8,-6.4) circle (1.4);
            \draw [black] (10.2,-3) -- (8.21,-4.99);
            \fill [black] (8.21,-4.99) -- (9.13,-4.77) -- (8.43,-4.07);
            \draw [black] (8.586,-5.518) arc (144:-144:1.5);
            \draw (11.8,-6.4) node [right] {$e_i:g_{e_i}$};
            \fill [black] (8.59,-7.28) -- (8.94,-8.16) -- (9.53,-7.35);
            \draw [black] (7.682,-8.186) arc (54:-234:1.5);
            \draw (6.8,-11.4) node [below] {$s_i:g_{s_i}$};
            \fill [black] (5.92,-8.19) -- (5.04,-8.54) -- (5.85,-9.13);
            \draw [black] (4.903,-7.009) arc (315.52567:27.52567:1.5);
            \draw (1.65,-5.55) node [left] {$w_i:g_{w_i}$};
            \fill [black] (5.16,-5.26) -- (4.94,-4.35) -- (4.24,-5.06);
            \draw [black] (5.918,-4.614) arc (234:-54:1.5);
            \draw (6.8,-1.4) node [above] {$n_i:g_{n_i}$};
            \fill [black] (7.68,-4.61) -- (8.56,-4.26) -- (7.75,-3.67);
        \end{tikzpicture}
    }
    \subfigure[{\G[6]}]{
        \begin{tikzpicture}[scale=0.2]
            \tikzstyle{every node}+=[inner sep=0pt]
            \draw [black] (2.2,-6.9) circle (2);
            \draw [black] (2.2,-6.9) circle (1.4);
            \draw [black] (5.6,-3.5) -- (3.61,-5.49);
            \fill [black] (3.61,-5.49) -- (4.53,-5.27) -- (3.83,-4.57);
            \draw [black] (3.082,-8.686) arc (54:-234:1.5);
            \draw (2.2,-11.9) node [below] {$out_i:g_{out_i}$};
            \fill [black] (1.32,-8.69) -- (0.44,-9.04) -- (1.25,-9.63);
            \draw [black] (1.318,-5.114) arc (234:-54:1.5);
            \draw (2.2,-1.9) node [above] {$in_i^{x,y}:g_{in_i^{x,y}}$};
            \fill [black] (3.08,-5.11) -- (3.96,-4.76) -- (3.15,-4.17);
        \end{tikzpicture}
    
    }
    \caption{%
        \G[5] and \G[6] model the restriction that a robot \R[i] at $(\x,\y)$ cannot go to a neighboring cell $(x,y)$ if $|h_{\x,\y} - h_{x,y}| > 1$.
		Guards $g_{e_i}, g_{w_i}, g_{n_i}$ and $g_{s_i}$ are given by $\bigvee \limits_{\forall (x,y)} \left( (x = \y \land y = \y) \land |h_{\x, \y} - h_{x,y} | \le 1\right)$, where $x$ and $x$  assume the following values: 
		a) for $g_{e_i}$, $x = \x$ and $y = \y+1$; 
		b) for $g_{w_i}$, $x = \x$ and $y = \y-1$; 
		c) for $g_{n_i}$, $x = \x - 1$ and $y = \y$; and, finally,
		d) for $g_{s_i}$, $x = \x + 1$ and $y = \y$.
		Guard $g_{in_i^{x,y}}$ is given by $(\x = 0 \land \y = 0) \land h_{x,y} \leq 1$, whereas $g_{out_i}$ is given by $\bigvee \limits_{\forall (x,y) \in \IO} \left( (\x = x \land \y = y) \land h_{\mathbf{x,y}} \le 1\right)$.
    }
    \label{img:Gclimb}
\end{figure}
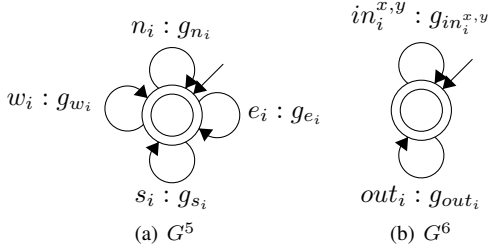

Given the climbing restriction of the robot, we develop \G[5] and \G[6] from \Fig~\ref{img:Gclimb} to prevent the robot $i$ from attempting to climb or go down steps greater than one.

As an example, let us consider that the robot is at location $(\x,\y)$ with height $h_{\x,\y} = 1$ and its neighbors $(\x + 1 \bcom \y)$ and $(\x \bcom \y+1)$ with $h_{(x+1),\y} = 3$ and $h_{\x,(\y+1)} = 2$. 
Then, in this case,  shifting to $(x + 1 \bcom \y)$ is avoided (event $e_i$), whereas for $(x \bcom \y + 1)$ it is allowed (event $s_i$).

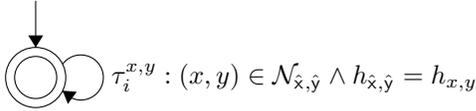
\begin{figure}[ht!]
    \centering
    \begin{tikzpicture}[scale=0.2]
        \tikzstyle{every node}+=[inner sep=0pt]
        \draw [black] (2.2,-5.3) circle (2);
        \draw [black] (2.2,-5.3) circle (1.4);
        \draw [black] (2.2,-0.2) -- (2.2,-3.3);
        \fill [black] (2.2,-3.3) -- (2.7,-2.5) -- (1.7,-2.5);
        \draw [black] (3.986,-4.418) arc (144:-144:1.5);
        \draw (7.2,-5.3) node [right] {$\evtT{x,y}{i}: (x,y) \in \N[\x,\y] \land h_{\x,\y} = h_{x,y}$};
        \fill [black] (3.99,-6.18) -- (4.34,-7.06) -- (4.93,-6.25);
    \end{tikzpicture}
    \caption{\G[7] models the restriction that the robot $i$ at $(\x \bcom \y)$ can unload a brick in one of its neighbors $(x \bcom y) \in \N[\x,\y]$ that is at the same height.}
    \label{img:Eadd}
\end{figure}

$\G[7]$, from \Fig~\ref{img:Eadd}, models the fact that a robot at $(\x \bcom \y)$ can only put a brick on a neighboring cell $(x \bcom y)\in \N[\x,\y]$ which has the same height as the cell where the robot is, \ie, $h_{\x,\y}=h_{x,y}$.

Finally, the behavior of robot \R[i] is given by the parallel composition $\R[i] = \G_3 \| \ldots \| \G_7$.
We note that \R[i] reaches a marked state whenever it is outside of the grid.

\subsection{Modular Synthesis}

Similar to the conventional supervisor synthesis process proposed by \cite{RW:1989}, our modular synthesis method starts from an automaton $\K[i] = \T[i] \| \R[i]$, which represents the desired behavior for \R[i] subject to structure \T[i], and computes a subautomaton of \K[i] that is co-accessible, \taskObs and \tDual.
Then, this subautomaton is used to implement \S[i].
These three properties, as well as our synthesis process, are formally described in \cite{MR:WODES2022}, in a more general context of coordination of multi-agent systems.
Here we describe the co-accessibility, \taskObs and \tDual properties in terms of the robotic construction problem as follows:

\begin{enumerate}
    \item[(P1)] \emph{co-accessibility}: from any intermediate structure \St[k] there is at least one sequence of brick laying that leads to target structure \St.
    This sequence may be performed collectively by \R[i] and other robots.
    Furthermore, when \St is achieved the robot \R[i] can leave the grid from its current position;
    \item[(P2)] \emph{\taskObs}: for any two sequential structures \St[k] and \St[k'], the robot \R[i] must always be in situations that
     \begin{enumerate}
         \item \emph{add brick}: it can move to some cell where it can attach the brick in \St[k] that leads to \St[k'];
         \item \emph{grant permission to other robots}: 
         if other robots can place the brick in \St[k] that leads to \St[k'], then \R[i] must be able to move to some position where it grants permission to other robots to do so;
     \end{enumerate}
    \item[(P3)] \emph{\tDual}: in a given intermediate structure \St[k], there is a situation where the robot \R[i] can place a brick on $(x,y)$ if and only if there is at least one situation where the robot \R[i] allows the other robots to place the brick on $(x,y)$.
\end{enumerate}

\subsection{Sufficient condition for solving the 3-D structure multi-robot construction problem}\label{sec:SufCond}

Now we state our sufficient condition for nonblocking of joint action of the supervisors.
This condition is stated in the context of the construction problem, but it is an instance of the more general case described in \cite{MR:WODES2022}.

\begin{theorem}[\cite{MR:WODES2022}]\label{thm:main}
    Given a set of construction robots $\{\R[1], \ldots, \R[n] \}$ and their respective supervisors $\{\S[1] \bcom \ldots \bcom \S[n]\}$, if \S[i], $i \in \{1, \ldots, n\}$, are nonblocking, \taskObs and \tDual, then the joint action of $\S[1], \ldots, \S[n]$ is nonblocking.
\end{theorem}

\subsection{Example}\label{sec:exCRC}

Now, we present an example to illustrate our approach.
In this example, we consider that the robots \R[1] and \R[2] must build the target structure \St from \Fig~\ref{img:strt}.

\begin{figure*}[ht!]
    \centering
    \includegraphics[width=\linewidth]{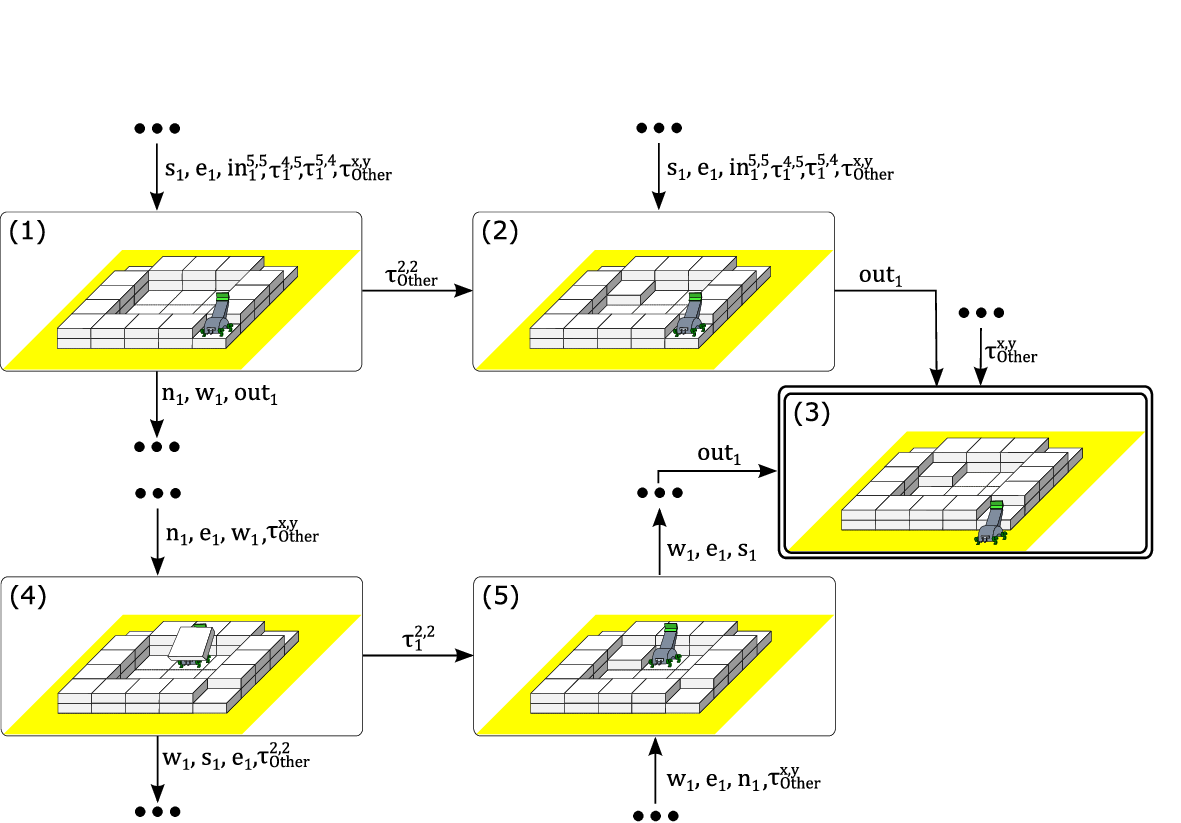}
    \caption{Part of automaton \K[1]. The three dots encapsulates sequences of states and transitions.  
    The arrows represent transitions, whereas the labels on them are events.}
    \label{img:Ki}
\end{figure*}

So, let \T[1] and \R[1] be constructed for the structure \St, with $\IO = \{(1,1), (1,5), (5,1), (5,5)\}$, as described in Sections~\ref{sec:structure} and \ref{sec:robot}.
In the example only the \taskObs property fails in $\K[1] = \T[1] \| \R[1]$.
It fails for the structure \St[k'] from state (1) in \Fig~\ref{img:Ki}.
This failure occurs due to the fact that when the robot \R[1] is in state (1), \R[1] cannot place the last brick on $(2,2)$.
The same happens for other states from \K[1] whose only difference is the current position of \R[1].
However, when the robot \R[1] is in state (4), then it can place a brick on $(2,2)$.
Thus, in this case our synthesis method prunes the state (1) and states similar to it from \K[1].
As a result, \R[1] is prohibited to attach the brick in all intermediate \St[k] that leads to \St[k'].
As the elimination of one state from \K[1] can alter the properties of another state, multiple iterations might be needed before the $\Synt(\K[1])$ procedure converge to a final automaton \S[1].
We point out that the states $(2), (3), (4)$ and $(5)$ survive in \S[1].

To obtain \S[2] from \S[1] we replace the index 1 with 2 from events $n_1, e_1, s_1, w_1, p_1, in_{1}^{x,y}, out_i$ and $\evtT{i}{x,y}$ of each transition in \S[1].
For instance, the events $n_1, e_1, s_1, w_1$ and \evtT{1}{2,2} from transitions connected to state (4) (\Fig~\ref{img:Ki}) are replace by $n_2, e_2, s_2, w_2$ and \evtT{2,2}{2}.

\begin{figure*}[ht!]
    \centering
    \includegraphics[width = \linewidth]{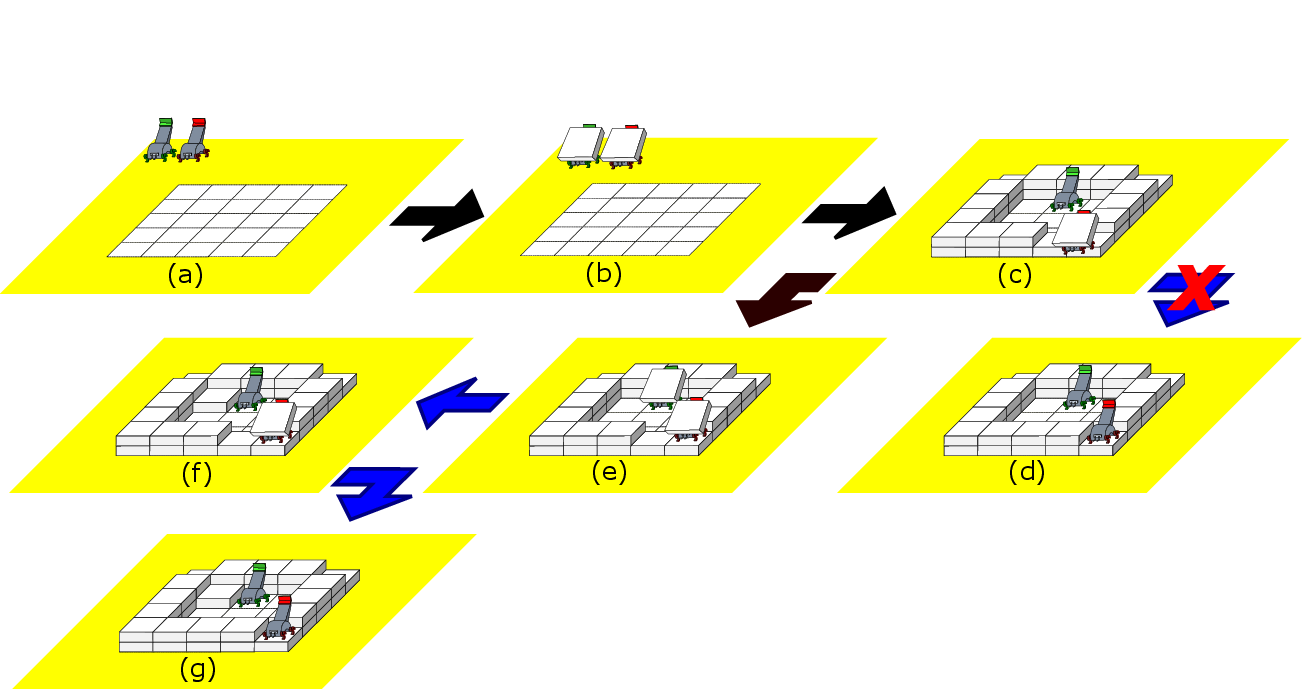}
    \caption{Construction of the target $\St$ with two construction robots, \R[1] (green robot) and \R[2] (red robot).
    Blue arrows indicate the occurrence of a single action, while black arrows indicate a sequence of actions.}
    \label{img:exSBS}
\end{figure*}

\Fig\ref{img:exSBS} illustrates the robots $\R[1]$ (green robot) and $\R[2]$ (red robot), under control of \S[1] and \S[2], working collectively to build the target structure $\St$.
Although in this example, for the sake of simplicity, only two robots are considered in the problem, our synthesis procedure is independent on the number of robots being used to build the same structure.
Next, we describe one of all possible sequences of intermediate structures that the two robots can build under control of $\S[1]$ and $\S[2]$. 

The construction process starts from structure $\St[0]$ with all robots situated outside of the grid and unloaded as shown in \Fig\ref{img:exSBS}~(a).
From this initial condition, \S[1] and \S[2] allow robots \R[1] and \R[2] to pick up a brick.
Assuming that both robots have picked-up bricks as in \Fig\ref{img:exSBS}~(b), then from this new situation \S[1] and \S[2] allow \R[1] and \R[2] to enter the grid through cells $(1,1), (1,5), (5,1)$ and $(5,5)$, or lay a brick on these cells.

Now, consider that the system evolves to the configuration presented in \Fig\ref{img:exSBS}~(c).
In this configuration, the supervisor $\S[2]$ prevents that robot \R[2] unloads a brick on $(5,4)$ until a brick is placed in $(2,2)$, because if it did, robot \R[1] would be stuck in the structure due to its climbing restriction.
Moreover, since both robots \R[1] and \R[2] cannot place the last brick on $(2,2)$, then the structure would not be completed.
Formally, this situation would correspond to a blocking state of the controlled system.

In our synthesis procedure $\S[1]$ and $\S[2]$ are obtained in such a way as to prevent these blocking states from being reached.
So, from situation illustrated in \Fig\ref{img:exSBS}~(c), \R[1] could move to $(0,0)$, pick up a brick and go back to $(2,3)$, whereas \R[2] waits at $(5,5)$ as in  \Fig\ref{img:exSBS}~(e).
Next, \R[1] could place a brick on $(2,2)$ and then \R[2] could place the second brick on $(5,4)$ as illustrated in  \Figs\ref{img:exSBS}~(f) and (g), respectively.


\section{CONCLUSION}\label{sec:conc}


We described in this paper a synthesis method based on SCT to solve a class of robotic construction problems.
In our approach, the synthesis process of a supervisor \S[i] for a robot \R[i] is local, since it only considers the models of robot \R[i] and structure \T[i].
It can be shown that the time complexity of this process is $O(N^2 + N^2M)$, where $N$ and $M$ are the numbers of states and transitions from \K[i], respectively.
It is worth noting that $N$ grows according to the size of the structure and, therefore, our synthesis method may face computational limitations with large structures.
However, whenever \S[i] may be synthesized, then our method is scalable in the sense that \S[i] can be replicated to any number of robots in a way that they are able to solve the construction problem. Besides, our solution offers flexibility due to the reactive and permissive nature of the control.

\addtolength{\textheight}{-12cm}   





\section*{ACKNOWLEDGMENT}

This work has been partially supported by CAPES, The Brazilian Agency for Higher Education, under the project Print CAPES-UFSC ``Automation 4.0''.

\bibliographystyle{IEEEtran}
\bibliography{Refs/Ref}

\end{document}